\documentclass[11pt,a4paper]{article}
\usepackage[utf8]{inputenc}
\usepackage[T1]{fontenc}
\usepackage{amsmath,amssymb}
\usepackage{booktabs}
\usepackage{graphicx}
\usepackage{hyperref}
\usepackage[margin=2.5cm]{geometry}
\usepackage{natbib}
\usepackage{xcolor}
\usepackage{flafter}

\hypersetup{
  colorlinks=true,
  linkcolor=blue!60!black,
  citecolor=blue!60!black,
  urlcolor=blue!60!black
}

\title{Cross-Entropy Is Load-Bearing:\\A Pre-Registered Scope Test of the K-Way Energy Probe\\on Bidirectional Predictive Coding}
\author{Jon-Paul Cacioli\\
Independent Researcher, Melbourne, Australia\\
ORCID: 0009-0000-7054-2014}
\date{April 2026}

\begin{document}
\maketitle

\begin{abstract}
\citet{cacioli2026ima} showed that the K-way energy probe on standard discriminative predictive coding networks reduces approximately to a monotone function of the log-softmax margin. The reduction rests on five assumptions, including cross-entropy (CE) at the output and effectively feedforward inference dynamics. This pre-registered study tests the reduction's sensitivity to CE removal using two conditions: standard PC trained with MSE instead of CE, and bidirectional PC (bPC; \citealt{oliviers2025bpc}).

Across 10 seeds on CIFAR-10 with a matched 2.1M-parameter backbone, we find three results. The negative result of \citet{cacioli2026ima} replicates on standard PC: the probe sits below softmax ($\Delta = -0.082$, $p < 10^{-6}$). On bPC the probe exceeds softmax across all 10 seeds ($\Delta = +0.008$, $p = 0.000027$), though a pre-registered manipulation check shows that bPC does not produce materially greater latent movement than standard PC at this scale (ratio 1.6, threshold 10). Removing CE alone without changing inference dynamics halves the probe-softmax gap ($\Delta_{\text{MSE}} = -0.037$ vs $\Delta_{\text{stdPC}} = -0.082$).

CE is a major empirically load-bearing component of the decomposition at this scale. CE training produces output logit norms approximately 15$\times$ larger than MSE or bPC training. A post-hoc temperature scaling ablation decomposes the probe-softmax gap into two components: approximately 66\% is attributable to logit-scale effects removable by temperature rescaling, and approximately 34\% reflects a scale-invariant ranking advantage of CE-trained representations. We use ``metacognitive'' operationally to denote Type-2 discrimination of a readout over its own Type-1 correctness, not to imply human-like introspective access.

\medskip
\noindent\textbf{Pre-registration:} \url{https://osf.io/2kvsp} (filed prior to data collection).\\
\textbf{Code and data:} \url{https://github.com/synthiumjp/ima}
\end{abstract}

\section{Introduction}

\subsection{The decomposition and its scope}

\citet{cacioli2026ima} established an approximate decomposition of the K-way energy probe on discriminative predictive coding networks \citep{rao1999,bogacz2017,whittington2017,millidge2022}. Under five assumptions (CE at the output (A1), target clamping (A2), effectively feedforward dynamics (A3), deterministic generative predictions (A4), encoder-generative consistency (A5)), the K-way energy margin decomposes as
\begin{equation}
M_k(x) \approx [\text{log-softmax margin}]_k + [R_{(2)}(x) - R_{(1)}(x)],
\end{equation}
where $R_k(x)$ is a residual from generative-chain propagation of the clamped target. This residual is not trained to correlate with correctness. The decomposition predicts that the probe should track softmax from below. This prediction was confirmed empirically across six conditions on a single seed.

\citet{cacioli2026ima} was explicit about scope. The decomposition does not apply to bidirectional PC, prospective configuration, generative PC at test time, or energy formulations without CE at the output. The present study tests the first and last of these exclusions.

\subsection{Bidirectional predictive coding}

\citet{oliviers2025bpc} introduced bidirectional PC (bPC). It replaces the standard discriminative PC energy with a bidirectional MSE formulation:
\begin{equation}
E_{\text{bPC}} = \sum_{l=1}^{L-1} \frac{\alpha_{\text{gen}}}{2} \overline{\|x_l - W_{l+1} f(x_{l+1})\|^2} + \sum_{l=2}^{L} \frac{\alpha_{\text{disc}}}{2} \overline{\|x_l - V_{l-1} f(x_{l-1})\|^2}.
\end{equation}
There is no CE term. Both top-down (generative, $W$) and bottom-up (discriminative, $V$) prediction errors drive latent updates during inference. The encoder and generative pathways are co-trained under a shared bidirectional energy objective.

Relative to the decomposition's assumptions, bPC violates A1 (no CE) and A3 (genuinely iterative bidirectional dynamics), and substantially alters A5. The decomposition's algebraic chain breaks at Step~1, where the CE term links settled energy to log-softmax. Whether the dynamics change (A3) or the energy-formulation change (A1) is the more consequential violation is the empirical question this study addresses. At the TinyConv scale tested here \citep{pinchetti2024}, bPC's generative chain contributes less than 1\% of total energy at the calibration-selected $\alpha_{\text{gen}}$. This constrains the regime in which results should be interpreted.

\subsection{This study}

We ask whether CE at the output (A1) is a load-bearing component of the decomposition. We compare three conditions on a matched backbone: standard discriminative PC with CE (baseline), standard PC with MSE replacing CE (isolating A1), and bidirectional PC (violating A1, A3, and altering A5 simultaneously).

The discPC-MSE condition violates only A1 while preserving A3 and A2. If it shows a smaller probe-softmax gap than standard PC, the energy formulation is partially responsible, independent of dynamics. The bPC condition was intended to additionally test A3. As reported below, the pre-registered manipulation check for dynamics failed. bPC did not produce materially greater latent movement than standard PC at this scale. The study therefore addresses A1 directly. It addresses A3 only negatively: the dynamics did not functionally manifest.

The study was pre-registered on the Open Science Framework before any training runs. All hypotheses, statistical tests, decision rules, and diagnostics were specified in advance. This paper reports the pre-registered analyses without modification. The temperature scaling ablation (\S\ref{sec:tempscaling}) was not pre-registered and is labelled as post-hoc throughout.

\section{Method}

\subsection{Architecture}

All three conditions use the same structural backbone (TinyConv, 2,144,938 parameters). The encoder/V-pathway consists of Conv~3$\to$32, BN, GELU, MaxPool, Conv~32$\to$64, BN, GELU, MaxPool, Linear~4096$\to$256 with GELU, and Linear~256$\to$10. The generative/W-pathway consists of Linear~10$\to$256, Linear~256$\to$4096 reshaped to 64$\times$8$\times$8, Interpolate(2) with Conv~64$\to$32. This matches the architecture used in \citet{cacioli2026ima}.

\subsection{Conditions}

\textbf{Condition A (stdPC-CE).} Standard discriminative PC with CE at the output. $T$=13 inference steps for training and evaluation. Data normalised to [0, 1] with channel-wise standardisation. This replicates the baseline of \citet{cacioli2026ima}.

\textbf{Condition B (stdPC-MSE).} Standard discriminative PC with MSE at the output instead of CE. All other settings identical to Condition~A. This isolates A1 from A3.

\textbf{Condition C (bPC).} Bidirectional PC following \citet{oliviers2025bpc}. Bidirectional MSE energy with no CE. $T$=32 for training, $T$=100 for evaluation. $\alpha_{\text{gen}}$ determined by a pre-registered 5-seed calibration sweep (selected value: $10^{-5}$). $\alpha_{\text{disc}} = 1.0$. Data normalised to [$-$1, 1].

All conditions use AdamW (lr=$10^{-4}$, weight decay=$10^{-4}$), batch size 128, 25 epochs.

\subsection{Evaluation}

For each trained network, two readouts are evaluated on the first 1,280 CIFAR-10 test images.

The \textbf{K-way energy probe} clamps each of $K$=10 candidate classes at the output, runs inference to settling, and records the per-hypothesis total energy. The structural prediction is argmin over classes. The confidence margin is $E_{(2)} - E_{(1)}$.

The \textbf{softmax baseline} reads the feedforward output logits from the V-pathway with no inference loop. The confidence margin is the top-1 minus top-2 softmax probability.

The primary outcome is $\Delta = \text{AUROC}_2^{\text{probe}} - \text{AUROC}_2^{\text{softmax}}$, computed within each network. AUROC$_2$ is Type-2 AUROC: the area under the ROC curve for discriminating correct from incorrect Type-1 predictions using the readout's confidence margin \citep{fleming2014metacognition,maniscalco2012sdt}.

\subsection{Design and hypotheses}

The design is paired within-seed: 10 seeds (6 to 15) $\times$ 3 conditions = 30 training runs. All conditions share the same seed for weight initialisation and data ordering.

\textbf{H3} (manipulation check, evaluated first). bPC latent movement exceeds stdPC by at least 10$\times$. Deterministic threshold.

\textbf{H2} (replication, evaluated second). Mean $\Delta_{\text{stdPC}} < 0$. One-sided one-sample $t$-test, $\alpha = 0.05$.

\textbf{H1} (primary test, evaluated third). Mean $D = \Delta_{\text{bPC}} - \Delta_{\text{stdPC}} > 0$. One-sided paired $t$-test, $\alpha = 0.05$.

\textbf{H4} (exploratory). Descriptive comparison of $\Delta_{\text{MSE}}$ to $\Delta_{\text{stdPC}}$.

H3 gates interpretability. H2 gates the baseline. H1 is the primary test. H4 is exploratory.

\section{Results}

\subsection{Calibration}

A pre-registered sweep over $\alpha_{\text{gen}} \in \{10^{-3}, 10^{-4}, 10^{-5}, 10^{-6}\}$ across 5 seeds found all four values within 1 percentage point of each other in Type-1 accuracy (77.9 to 78.1\%). The tie-break rule selected $\alpha_{\text{gen}} = 10^{-5}$. Energy margins were invariant to $\alpha_{\text{gen}}$ across three orders of magnitude ($\sim$1.5$\times 10^{-4}$). The generative chain does not contribute meaningfully to hypothesis discrimination at any tested weighting.

\subsection{Main results}

Table~\ref{tab:main} shows condition means across 10 seeds.

\begin{table}[htbp]
\centering
\caption{Condition means ($N$ = 10 seeds).}
\label{tab:main}
\begin{tabular}{lccccc}
\toprule
Condition & Soft.\ Acc & Probe Acc & Soft.\ AUROC$_2$ & Probe AUROC$_2$ & $\Delta$ \\
\midrule
stdPC-CE  & 78.5\% & 69.2\% & 0.842 & 0.760 & $-$0.082 \\
stdPC-MSE & 79.0\% & 71.1\% & 0.836 & 0.799 & $-$0.037 \\
bPC       & 77.4\% & 77.6\% & 0.824 & 0.832 & $+$0.008 \\
\bottomrule
\end{tabular}
\end{table}

\subsection{H3: manipulation check, not confirmed}

The maximum per-layer latent movement ratio (bPC / stdPC) ranged from 1.53 to 1.67 across seeds. This is well below the pre-registered threshold of 10. bPC does not exhibit materially greater settled-state displacement than standard PC at this scale. Movement is concentrated in Layer~3, the 256-dimensional FC layer. Convolutional layers 1 and 2 show effectively zero displacement in all conditions.

The criterion for a strong dynamics manipulation was not met. The results do not support a dynamics-based interpretation under the planned standard. All subsequent results are reported with this scope limitation. bPC serves in the remainder of this paper as an independent CE-removal condition rather than as a test of bidirectional inference.

\subsection{H2: replication, confirmed}

Mean $\Delta_{\text{stdPC}} = -0.082$ (SD = 0.016). One-sided one-sample $t$-test: $t(9) = -16.0$, $p < 10^{-6}$, Cohen's $d = -5.06$. 95\% CI: [$-$0.094, $-$0.070]. All 10 seeds negative. Shapiro-Wilk: $W = 0.97$, $p = 0.90$. Wilcoxon: $W = 0$, $p = 0.001$.

The negative result of \citet{cacioli2026ima} replicates under multi-seed conditions. The K-way energy probe sits below softmax on standard discriminative PC, as predicted by the decomposition.

\subsection{H1: primary test, confirmed with scope limitation}

Mean $D = \Delta_{\text{bPC}} - \Delta_{\text{stdPC}} = +0.090$ (SD = 0.015). One-sided paired $t$-test: $t(9) = 19.2$, $p < 10^{-7}$, $d_z = 6.07$. 95\% CI: [0.079, 0.100]. All 10 seeds positive. Shapiro-Wilk: $W = 0.96$, $p = 0.83$. Wilcoxon: $W = 55$, $p = 0.001$.

Supplementary test: mean $\Delta_{\text{bPC}} = +0.008$ (SD = 0.004). One-sided one-sample $t$-test: $t(9) = 7.15$, $p = 0.000027$, $d = 2.26$. 95\% CI: [0.005, 0.010]. All 10 seeds positive.

The probe exceeds softmax on bPC across all seeds. H3 failure means this cannot be attributed to iterative dynamics. The result is better understood as a consequence of CE removal, which bPC shares with the MSE condition, rather than as an architectural effect of bidirectional inference.

\subsection{Softmax baseline validity, valid}

Both pre-registered triggers were checked. bPC probe accuracy exceeds bPC softmax accuracy by 0.15 percentage points (threshold: 5pp). stdPC softmax AUROC$_2$ exceeds bPC softmax AUROC$_2$ by 0.018 (threshold: 0.05). The primary comparison is diagnostic.

\subsection{H4: exploratory, A1 is partially load-bearing}

Mean $\Delta_{\text{MSE}} = -0.037$ (SD = 0.014). The within-seed difference $\Delta_{\text{MSE}} - \Delta_{\text{stdPC}} = +0.045$ (SD = 0.020). Exploratory paired $t$-test: $t(9) = 7.09$, $p = 0.000029$, $d_z = 2.24$. All 10 seeds show $\Delta_{\text{MSE}} > \Delta_{\text{stdPC}}$.

Removing CE alone reduces the probe-softmax gap from 0.082 to 0.037, a 54\% reduction. The energy formulation (A1) is a major empirically load-bearing component of the reduction at this scale. The remaining gap ($-$0.037) is substantial. A1 does not fully explain the reduction. Residual contributions from the asymmetric encoder-decoder training relationship (A5) or the generative chain mismatch likely account for the remainder. The present design suggests partial rather than exhaustive explanation.

\begin{figure}[htbp]
\centering
\includegraphics[width=0.85\textwidth]{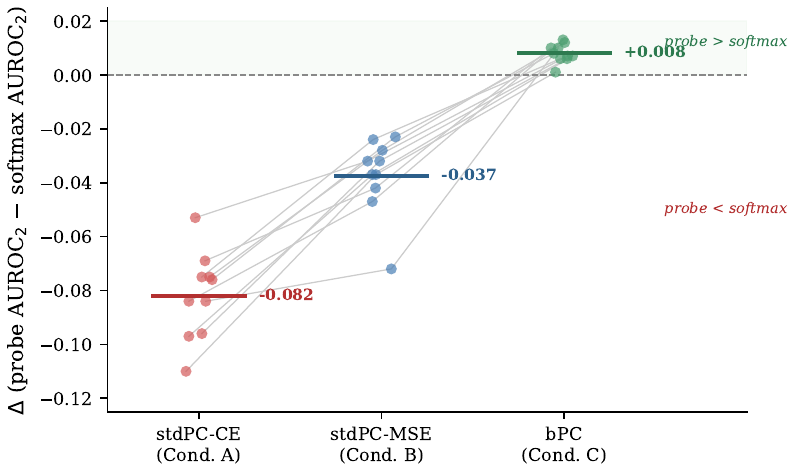}
\caption{Probe-softmax gap ($\Delta$) across three training conditions. Each point is one seed. Lines connect matched seeds across conditions. Horizontal bars show condition means. All 10 bPC seeds are above zero. All 10 stdPC-CE seeds are below.}
\label{fig:delta}
\end{figure}

Figure~\ref{fig:delta} shows the three-condition gradient with per-seed paired values. The monotonic shift from stdPC-CE through stdPC-MSE to bPC is visible in every seed pair.

\section{Logit Scale and the Probe-Softmax Gap}

The three-condition gradient in Table~\ref{tab:main} is associated with a systematic difference in output logit scale.

\subsection{CE training produces large logit norms}

CE training on standard PC produces output logit norms of approximately 11.5 (Table~\ref{tab:logits}). MSE and bPC training produce norms of approximately 0.78, a factor of 15$\times$ smaller.

\begin{table}[htbp]
\centering
\caption{Logit diagnostics (means across 10 seeds).}
\label{tab:logits}
\begin{tabular}{lccc}
\toprule
Condition & Logit norm & Logit margin & Softmax AUROC$_2$ \\
\midrule
stdPC-CE  & 11.52 & 2.69 & 0.842 \\
stdPC-MSE & 0.78  & 0.49 & 0.836 \\
bPC       & 0.78  & 0.45 & 0.824 \\
\bottomrule
\end{tabular}
\end{table}

\begin{figure}[htbp]
\centering
\includegraphics[width=0.65\textwidth]{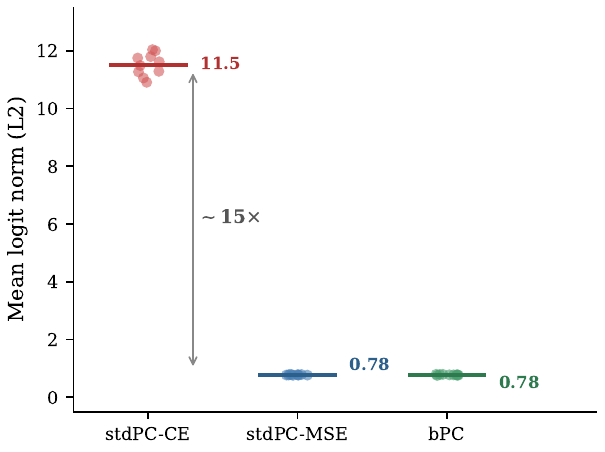}
\caption{Output logit norms (L2) across training conditions. Each point is one seed. CE-trained models produce logit norms approximately 15$\times$ larger than MSE or bPC.}
\label{fig:logitnorms}
\end{figure}

The CE gradient is proportional to $(\hat{p}_k - y_k)$, which is always non-zero for non-target classes. The optimiser satisfies CE by making the correct-class logit large relative to others. MSE pushes the output toward the one-hot target without incentivising large logit magnitudes beyond matching the target. bPC has no output-layer loss at all. Logit scale is determined by the encoder's internal dynamics. Figure~\ref{fig:logitnorms} shows the per-seed logit norms across conditions.

\subsection{Large logits are associated with higher softmax AUROC$_2$}

When logit norms are large, the softmax distribution sharpens. The maximum softmax probability approaches 1.0 for both correct and incorrect predictions. AUROC$_2$ is invariant to monotone transformations of the scalar confidence score \citep{pepe2003}. Softmax is not a monotone function of any single logit. It is a nonlinear function of the full logit vector. Changes in logit scale can therefore change the ranking structure of softmax confidence margins across correct and incorrect predictions, which can change AUROC$_2$.

CE-trained models show both larger logit norms and higher softmax AUROC$_2$ ($\sim$0.84 vs $\sim$0.82 to 0.83). The energy probe operates on prediction errors in the generative chain, not on output logits. It is not directly affected by logit scale. When softmax is favoured by CE-linked logit scaling, the energy probe appears to underperform. This may not reflect weakness in the probe. It may reflect the baseline being favoured by the training objective.

\subsection{Removing CE removes the logit-scale difference}

When CE is replaced by MSE (Condition~B), logit norms drop from 11.5 to 0.78 and the probe-softmax gap halves. When the entire energy formulation is replaced by bidirectional MSE (Condition~C), logit norms are similarly small and the probe slightly exceeds softmax. The association between logit scale and the probe-softmax gap is monotonic across the three conditions. The cross-condition comparison alone does not fully isolate this as the causal mechanism.

\subsection{Direct test: temperature scaling on CE-trained models}
\label{sec:tempscaling}

To test the logit-scale account directly, we applied post-hoc temperature scaling to the 10 stdPC-CE models. For each seed, we divided the raw output logits by a temperature $T$ chosen to approximately match the mean logit norm across the test set to the MSE condition ($\sim$0.78), yielding $T \approx 15$. We then recomputed softmax AUROC$_2$ on the rescaled logits. This changes the softmax probability distribution without retraining or altering the learned representations.

AUROC$_2$ is invariant to monotone transformations of the scalar confidence score \citep{pepe2003}. Temperature scaling is a monotone transformation of the logit vector. The softmax confidence margin is a nonlinear function of the full logit vector, not a monotone function of any individual logit. Temperature scaling therefore changes the confidence margin's ranking across examples. This is not a violation of the invariance theorem. It is a consequence of the score function itself being transformed.

\begin{table}[htbp]
\centering
\caption{Temperature scaling ablation (means across 10 seeds).}
\label{tab:tempscale}
\begin{tabular}{lc}
\toprule
Metric & Value \\
\midrule
Softmax AUROC$_2$ (original, $T$=1) & 0.841 \\
Softmax AUROC$_2$ (rescaled, $T \approx$15) & 0.789 \\
Probe AUROC$_2$ & 0.761 \\
$\Delta$ (original) & $-$0.080 \\
$\Delta$ (rescaled) & $-$0.027 \\
Gap reduction & 0.053 (66\%) \\
\bottomrule
\end{tabular}
\end{table}

Temperature scaling reduced the probe-softmax gap by 66\% (Table~\ref{tab:tempscale}). The remaining gap ($-$0.027) persists even at extreme temperatures. A sweep from $T$=1 to $T$=30 (Table~\ref{tab:tempsweep}) shows softmax AUROC$_2$ plateauing at approximately 0.784, still above the probe's 0.761. The qualitative split is robust: the gap decreases rapidly between $T$=1 and $T$=10, then asymptotes.

\begin{table}[htbp]
\centering
\caption{Temperature sweep (means across 10 seeds).}
\label{tab:tempsweep}
\begin{tabular}{ccc}
\toprule
$T$ & Softmax AUROC$_2$ & Logit norm \\
\midrule
1.0  & 0.841 & 11.52 \\
2.0  & 0.830 & 5.76 \\
5.0  & 0.805 & 2.30 \\
10.0 & 0.793 & 1.15 \\
15.0 & 0.789 & 0.77 \\
20.0 & 0.786 & 0.58 \\
30.0 & 0.784 & 0.38 \\
\bottomrule
\end{tabular}
\end{table}

These results suggest a descriptive decomposition of the probe-softmax gap into two components. Temperature scaling alters the softmax mapping in ways that may affect both scale and ranking simultaneously. The partition should be interpreted as an empirical approximation rather than a strict causal separation.

\textbf{Component 1, scale-dependent ($\sim$66\%).} CE-linked logit inflation changes how softmax distributes confidence across correct and incorrect predictions. Temperature scaling removes this. It accounts for approximately 0.053 of the original 0.080 gap.

\textbf{Component 2, scale-invariant ($\sim$34\%).} CE training produces a classifier whose logit ordering is better correlated with correctness than the energy probe's energy ordering, independent of scale. This persists at any temperature and accounts for approximately 0.027 of the gap. CE is a strictly proper scoring rule that directly optimises the log-probability of the correct class \citep{gneiting2007}. This induces representations whose logit ordering approximates likelihood ratios. It provides a principled explanation for why CE-trained softmax retains a ranking advantage after scale correction.

\begin{figure}[htbp]
\centering
\includegraphics[width=0.85\textwidth]{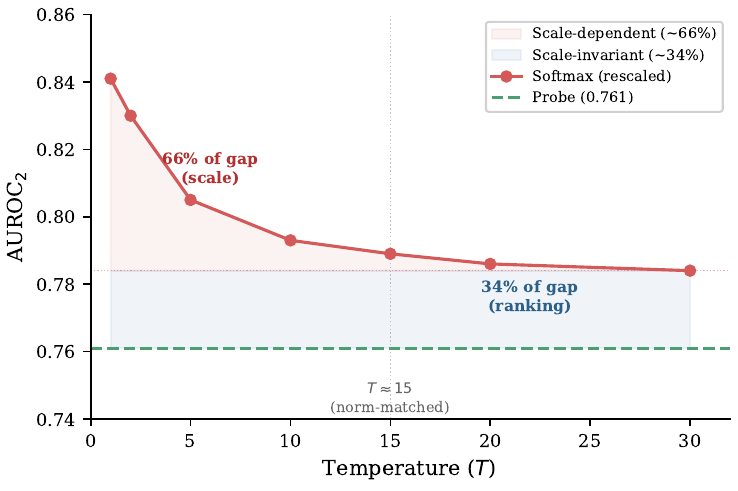}
\caption{Temperature scaling decomposes the probe-softmax gap. Softmax AUROC$_2$ on stdPC-CE models at varying temperature $T$ (means across 10 seeds). The dashed green line shows probe AUROC$_2$. Approximately 66\% of the gap is attributable to scale (removable by rescaling) and 34\% persists as a scale-invariant ranking advantage. The vertical dotted line marks $T \approx 15$, the norm-matched point.}
\label{fig:tempsweep}
\end{figure}

Figure~\ref{fig:tempsweep} shows the full temperature sweep. The softmax AUROC$_2$ curve drops steeply between $T$=1 and $T$=10, then plateaus. The shaded regions illustrate the two components of the gap.

\section{Energy Decomposition}

For Condition~C (bPC), the per-hypothesis settled energy was decomposed into generative and discriminative components.

The generative contribution averaged 0.000002 across seeds and hypotheses. The discriminative contribution averaged 0.000255. The gen/disc ratio was approximately 0.009. At $\alpha_{\text{gen}} = 10^{-5}$, the generative chain contributes less than 1\% of total energy. The K-way probe on bPC is dominated by discriminative (V-pathway) prediction errors.

The probe's advantage on bPC does not arise from the generative chain providing additional hypothesis-discriminating information. It is better explained by the absence of CE-linked logit inflation in the softmax baseline than by the presence of a richer energy landscape.

\section{Discussion}

\subsection{What the decomposition depends on}

The present data indicate that the reduction is more sensitive to A1 (CE at the output) than to A3 (inference dynamics) in the regime tested. Removing CE alone halves the probe-softmax gap. Adding bidirectional dynamics on top of CE removal tips the balance to a small positive Delta. The dynamics contribution is not clearly separable from the CE-removal effect. The manipulation check fails, the generative chain contributes less than 1\% of energy, and the three-condition gradient tracks logit scale rather than showing a step function at the dynamics boundary.

The theoretical point is clear. Outside A1, the original monotone-link derivation no longer applies. The CE term links settled energy to log-softmax. Remove CE and the energy at the output becomes MSE between the output latent and the clamped one-hot, which is not a monotone function of the log-softmax probability. The empirical point is that in the present experiments, removing CE is associated with smaller logits and a better relative showing for the probe. Whether these two points are connected by logit-scale inflation specifically is addressed by the temperature ablation, which confirms it as a major contributor (66\%) but not the sole mechanism.

\subsection{The logit-scale mechanism: confirmed but partial}

The temperature scaling ablation (\S\ref{sec:tempscaling}) confirms that logit-scale inflation is a major contributor, accounting for approximately two-thirds of the original $\Delta$ on CE-trained models. The remaining third represents a scale-invariant ranking advantage attributable to CE's role as a proper scoring rule \citep{gneiting2007,guo2017,wei2022}.

Softmax AUROC$_2$ on CE-trained networks is not a neutral baseline. Approximately two-thirds of its advantage is removable by post-hoc temperature scaling. Comparisons between structural probes and softmax should consider temperature-scaled softmax as a fairer reference point.

\subsection{What is not established}

Several limitations constrain the interpretation.

The manipulation check failed. bPC did not produce meaningfully greater latent movement than standard PC. The H1 result cannot be attributed to bidirectional dynamics. A clean test of A3 alone would require prospective configuration \citep{song2024} or a discPC model with genuinely non-trivial inference. Neither was tested.

The absolute magnitude of $\Delta_{\text{bPC}}$ is small (+0.008). The probe exceeds softmax, but barely. Both readouts produce similar AUROC$_2$ values ($\sim$0.82 to 0.84) on CIFAR-10 at this scale.

All experiments use a single architecture (TinyConv, 2.1M parameters, 4 effective layers) on a single dataset (CIFAR-10). The generative chain has 3 layers and contributes less than 1\% of energy at $\alpha_{\text{gen}} = 10^{-5}$. At deeper architectures or higher $\alpha_{\text{gen}}$, the generative chain might contribute more and the results might differ \citep{innocenti2025,goemaere2025}.

The study tests bPC in its supervised classification regime only. \citet{oliviers2025bpc} demonstrate bPC's advantages in multimodal learning and missing-input inference, where bidirectional dynamics are more functionally consequential.

\subsection{Relation to the broader programme}

This study is part of a programme on metacognitive measurement in neural networks. Prior work documented failure modes of single-point confidence probes on frontier language models, including RLHF as a response-set inducer that dominates confidence signals. \citet{cacioli2026ima} proposed structural energy-landscape probing as an alternative and showed that it does not escape the softmax baseline on standard discriminative PC.

The present finding adds a mechanistic dimension. The observed probe-softmax disadvantage on CE-trained networks is associated with CE-linked logit scaling rather than with a fundamental limitation of structural probes. How the network is trained to express confidence appears to matter at least as much as how the network is architecturally structured, though the present study establishes this as an empirical pattern rather than a confirmed causal mechanism.

\section{Conclusion}

In this pre-registered scope test, CE at the output emerges as the major empirically load-bearing component of the decomposition at TinyConv scale. Removing CE allows the K-way energy probe to match or exceed softmax. This holds whether CE is replaced by MSE or by bPC's bidirectional MSE formulation.

A post-hoc temperature scaling ablation decomposes CE's contribution into two channels. Approximately 66\% of the probe-softmax gap is attributable to logit-scale inflation. CE training produces logit norms approximately 15$\times$ larger than MSE or bPC, and rescaling logits to match MSE norms closes two-thirds of the gap. The remaining 34\% reflects a genuine ranking advantage. CE-trained representations produce softmax confidence signals that are better correlated with correctness, independent of scale. CE does not merely inflate the baseline. It also produces a better-ranked baseline.

Bidirectional dynamics are not the operative mechanism at this scale. The pre-registered manipulation check failed. bPC latent movement was 1.6$\times$ standard PC movement, well below the 10$\times$ threshold. The generative chain contributed less than 1\% of the K-way probe's energy.

The more consequential finding is the three-condition gradient itself. The probe is no longer structurally dominated once CE is removed. This suggests that the negative result of \citet{cacioli2026ima} is correctly scoped to CE-trained discriminative PC and should not be read as a general limitation of structural probing. For confidence probe evaluation more broadly, temperature-scaled softmax may be a fairer baseline than raw softmax when comparing alternative confidence signals on CE-trained networks.

\section*{Acknowledgments}

This research is self-funded. The author has no external funding, institutional affiliation, or conflicts of interest. Computational experiments were conducted on a consumer GPU (AMD Radeon RX 7900 GRE, 16~GB VRAM).

\end{document}